\documentclass[sigconf]{acmart}
\usepackage{subcaption}
\usepackage{enumitem}
\usepackage{booktabs}
\usepackage{pifont}
\usepackage{multirow}
\usepackage{makecell}
\usepackage{algorithm}
\usepackage{algpseudocode}
\usepackage{placeins}
\usepackage{float}
\usepackage{xcolor}
\usepackage{dblfloatfix}

\AtBeginDocument{%
  }

\setcopyright{acmlicensed}
\copyrightyear{2018}
\acmYear{2018}
\acmDOI{XXXXXXX.XXXXXXX}
\acmConference[Conference acronym 'XX]{Make sure to enter the correct
  conference title from your rights confirmation email}{June 03--05,
  2018}{Woodstock, NY}
\acmISBN{978-1-4503-XXXX-X/2018/06}

\begin{document}

\title{FLUID: From Ephemeral IDs to Multimodal Semantic Codes for Industrial-Scale Livestreaming Recommendation}

\author{Xinhang Yuan}
\authornote{Core contributors.}
\email{rose.yuan@tiktok.com}
\affiliation{%
  \institution{TikTok}
  \city{San Jose}
  \state{California}
  \country{USA}
}
\author{Zexi Huang}
\authornotemark[1]
\authornote{Corresponding author.}
\email{zexi.huang@tiktok.com}
\affiliation{%
  \institution{TikTok}
  \city{San Jose}
  \state{California}
  \country{USA}
}
\author{Anjia Cao}
\authornotemark[1]
\email{caoanjia@bytedance.com}
\affiliation{%
  \institution{ByteDance}
  \city{Shanghai}
  \country{China}
}
\author{Xudong Lu}
\authornotemark[1]
\email{luka.30@bytedance.com}
\affiliation{%
  \institution{ByteDance}
  \city{Shanghai}
  \country{China}
}

\author{Zikai Wang}
\email{wangzikai.kevin@bytedance.com}
\affiliation{%
  \institution{ByteDance}
  \city{San Jose}
  \state{California}
  \country{USA}
}

\author{Penghao Zhou}
\email{zhoupenghao.leon@bytedance.com}
\affiliation{%
  \institution{ByteDance}
  \city{Shanghai}
  \country{China}
}
\author{Chang Liu}
\email{chang.liu.8@bytedance.com}
\affiliation{%
  \institution{ByteDance}
  \country{Singapore}
}
\author{Wentao Guo}
\email{wentao.guo@bytedance.com}
\affiliation{%
  \institution{ByteDance}
  \city{San Jose}
  \state{California}
  \country{USA}
}

\author{Qinglei Wang}
\email{wangqinglei@bytedance.com}
\affiliation{%
  \institution{ByteDance}
  \country{Singapore}
}

\renewcommand{\shortauthors}{Yuan$^*$, Huang$^*$, Cao$^*$, Lu$^*$ et al.}


\begin{abstract}
Modern recommender systems rely heavily on ID-based collaborative filtering: each item is represented by a unique ID embedding that accumulates collaborative signals from user interactions. Livestreaming recommendation, however, faces a unique challenge in this paradigm: a live room typically broadcasts for only tens of minutes, so its item ID remains poorly learned in a persistent cold-start state and ID-centric ranking models fail to generalize. We present FLUID, the first framework to fully retire the candidate-side item ID from a production-scale livestreaming ranker. FLUID introduces a cross-domain multimodal encoder, jointly trained on short videos and livestreams, to produce discrete hierarchical semantic codes, called LUCID, for content-based item characterization. To adapt the ranker to LUCID, FLUID further employs a staged warmup scheme: it first incorporates cold, slice-level LUCID as an independent token alongside the ID embedding, and then replaces the ID embedding with warm, room-level LUCID before online incremental training. Deployed on our industrial livestreaming recommenders with a cross-platform combined user base of over one billion globally, FLUID delivers significant online gains of +0.55\% Quality Watch Duration, +2.05\% Cold-Start Room Views, and +0.05\% Active Hours.
\end{abstract}







\maketitle

\section{Introduction}

\begin{figure}[!htbp]
    \centering
    \includegraphics[width=0.95\linewidth]{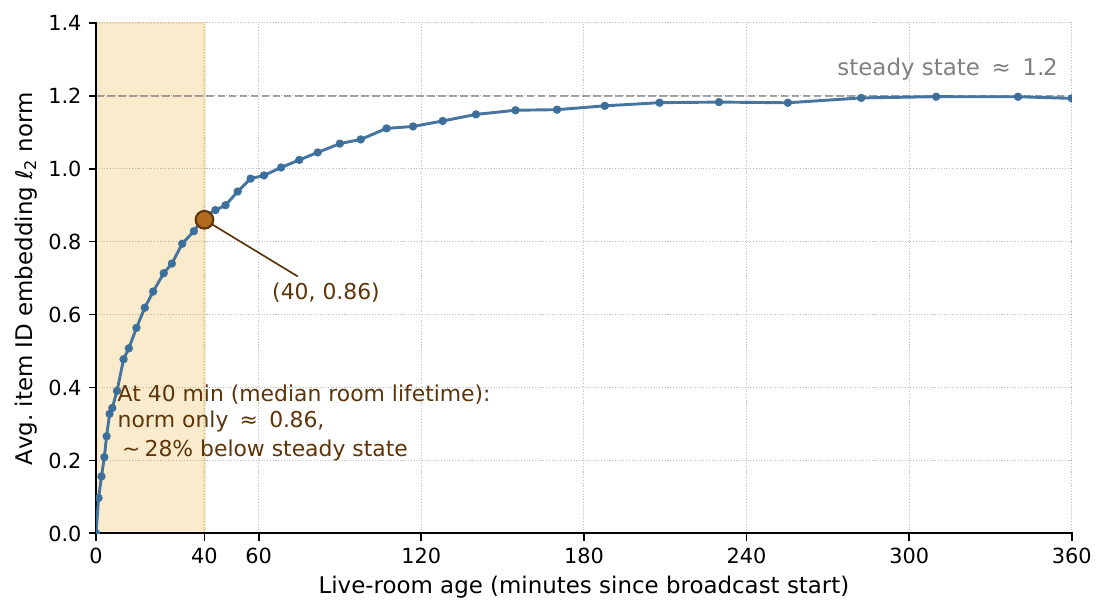}
    \caption{Item ID embedding $\ell_2$ norm vs.\ live-room age, aggregated over one day of production traffic. The norm fails to converge within the median ${\sim}40$-minute room lifetime.}
    \label{fig:gid_norm_vs_age}
\end{figure}

Livestreaming has emerged as one of the fastest-growing online content ecosystems~\cite{zhang2026zenith,liu2025larm,yang2026sarm}, where creators broadcast in real time and users interact through viewing, commenting, and gifting. Unlike videos (typical item lifetime measured in days to months) and e-commerce (months to years), livestreams are only relevant when on air. On our leading livestreaming platforms with a combined user base of over one billion, a live room typically broadcasts for tens of minutes (median ${\sim}40$\,min). This poses a fundamental challenge for ID-centric recommender systems, which rely on accumulating collaborative signals on item ID embeddings to power personalization. As shown in ~\autoref{fig:gid_norm_vs_age}, the embedding norm fails to converge within the median 40-minute room lifetime. Most items therefore spend their entire lifetime with undertrained embeddings. Meanwhile, modern large recommendation models (LRMs)~\cite{zhai2024actions,zhang2024wukong,gui2023hiformer,zhang2026zenith} still place most of their capacity in the ID embedding tables, which only memorize short-lived exposure signals and struggle to generalize in the livestreaming setting. 

A promising solution to reduce ranker reliance on memorization-based ID embeddings is to leverage content-derived signals. One line of work appends frozen multimodal embeddings to items as auxiliary features~\cite{sheng2024taobao,luo2025qarm}. Another aligns or discretizes multimodal representations into compact semantic codes with learnable embedding tables similar to ID features~\cite{rajput2023recommender, liu2024alignrec,luo2025qarm,ye2025dual,deng2024em3,abrec2025,zheng2025semid}. 
Further, end-to-end multimodal recommenders~\cite{liu2025larm,zhang2024notellm2,yang2026sarm} train the multimodal encoder jointly with downstream ranking tasks. Despite their diverse forms, all these systems treat multimodal features as \emph{complementary} signals \emph{alongside} the dominant item IDs, including those specifically designed for livestreaming ~\cite{liu2025larm}. Yet, two livestreaming-specific challenges remain unsolved.

\emph{First, producing high-quality semantic representations for livestreaming is nontrivial.} Live content is inherently multimodal and fast-changing, with visuals, speech, on-screen text, streamer metadata, and audience signal all evolve within minutes~\cite{liu2025larm,yang2026sarm}. Content representations must capture both transient dynamics and persistent room characteristics. In addition, unlike videos and e-commerce where contents are themed and user feedback signals are dense, which majority of multimodal encoders are designed for ~\cite{luo2025qarm,deng2024em3,sheng2024taobao,rajput2023recommender}, livestreaming can be more challenging to clearly characterize with less supervised signals for encoders to train on. 

\emph{Second, incorporating semantic representations into ID-centric rankers is difficult.} When adding multimodal features on top of ID-based signals, rankers often take the ``shortcut'' of ID-embeddings and underutilize the multimodal input, usually requiring explicit alignment, optimization rebalancing, or contrastive regularization to strengthen multimodal contribution~\cite{luo2025qarm,deng2024em3,abrec2025}. This ID-dominant effect is especially critical to livestreaming as item IDs are ephemeral and do not carry as much collaborative information as in other content recommendation scenarios. 

\begin{figure*}[!htbp]
    \centering
    \includegraphics[width=\textwidth]{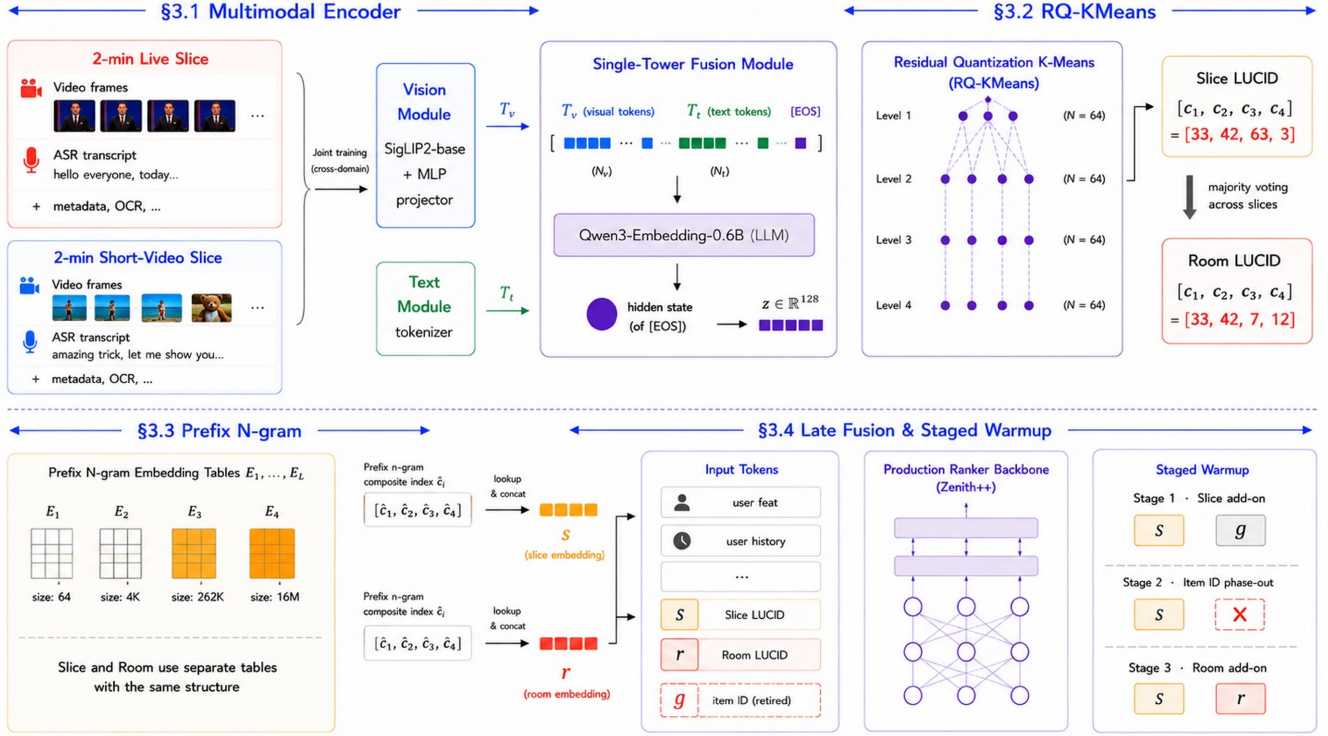}
    \caption{Overview of FLUID. \textit{Top:} a cross-domain multimodal encoder (SigLIP2 ViT + Qwen3-Embedding) jointly trained on livestreams and short videos produces a 128-d slice embedding $z$, which RQ-KMeans discretizes into a 4-level codeword tuple---\emph{LUCID}---with room-level LUCID obtained by per-level majority voting over slices in a session. \textit{Bottom:} slice-level and room-level LUCID enter the token-based ranker backbone as independent tokens (late fusion) via prefix n-gram embeddings, replacing the item id embedding with a staged warmup scheme.}
    \label{fig:framework_overview}
\end{figure*}

To address these two challenges, we propose \textbf{FLUID} (\textbf{F}ramework for \textbf{L}ive \textbf{U}niversal \textbf{ID}-free Recommendation, \autoref{fig:framework_overview}), the first framework to \emph{fully replace} item ID with content-based semantic codes in industrial-scale livestreaming rankers. Specifically, FLUID first introduces a \emph{cross-domain} multimodal encoder jointly trained on short videos and livestreams to leverage clean and dense supervision signals from the short-video domain, producing discrete semantic codes we call \textbf{LUCID} (\textbf{L}ive \textbf{U}niversal \textbf{C}ontent \textbf{ID}entifier). Second, a \emph{staged warmup} training scheme adapts the ranker to LUCID by first leveraging the ranker backbone for a late fusion of cold, slice-level LUCID embedding and the existing item ID embedding and then replacing the item ID embedding with warm, room-level LUCID embedding before the final online incremental training.
 
In summary, our contributions are as follows:
\begin{itemize}
    \item \textbf{Problem framing.} We argue that when the item ID is ephemeral, the ID-dominance effect becomes a fundamental bottleneck rather than a tolerable nuisance, motivating full retirement of the item ID for optimized ranking performance.
    \item \textbf{Cross-domain semantic representations.} A multimodal encoder jointly trained on short-video and livestream supervision produces discrete hierarchical codes that converts to embeddings via the prefix-n-gram scheme. 
    \item \textbf{Late-fusion ID-free ranker.} We deploy the first production-scale livestreaming ranker without item IDs thanks to our staged warmup scheme, which adds LUCID and replaces the item ID with a late fusion strategy. 
    
\end{itemize}
Deployed on our industrial livestreaming recommenders with a combined user base of over one billion globally, FLUID delivers significant online gains of $+0.55\%$ Quality Watch Duration, $+2.05\%$ Cold-Start Room Views, $+2.87\%$ Niche Room Views, $+1.63\%$ Unique Watched Tags, and $+0.05\%$ Active Hours.

\section{Related Work}
\label{sec:relatedwork}


\subsection{Multimodal Embedding}

Multimodal embedding learning aligns heterogeneous data---such as images, text, and videos---into a unified semantic space~\cite{radford2021clip,jia2021align}, enabling diverse downstream applications including industrial recommendation~\cite{liu2025larm,luo2025qarm}. Early work pioneered by CLIP~\cite{radford2021clip} adopts a \emph{dual-tower} paradigm with independent unimodal encoders aligned via contrastive learning, and has been extensively refined in specific aspects such as training objectives, architecture and applications~\cite{li2022blip,yu2022coca,zhai2023siglip,tschannen2025siglip2,zhang2024long-clip,cao2025flame,huang2024llm2clip,tao2022self,deng2024em3}. More recent works repurpose \emph{single-tower} paradigms for deeper cross-modal fusion, leveraging the rich world knowledge and reasoning capacity of pre-trained (multimodal) large language models~\cite{jiang2024vlm2vec,zhang2024gme,meng2025vlm2vecv2,chen2025moca,xiao2025metaembed,zhang2024notellm2,liu2025larm,liu2024alignrec}. Our multimodal encoder follows this \emph{single-tower} line.

Prior multimodal encoders for recommendation are also typically trained on data from a \emph{single domain}, such as live streaming, short videos, or e-commerce~\cite{liu2025larm,luo2025qarm,deng2024em3,sheng2024taobao,rajput2023recommender}. In contrast, our multimodal encoder is trained on \emph{cross-domain} data, enabling it to capture more robust and transferable semantic signals across diverse scenarios.

\subsection{ID-based Recommendation Models}

The dominant paradigm in modern recommender systems represents users and items as ID embeddings that carry collaborative signals. From early collaborative filtering~\cite{sarwar2001item} and matrix factorization~\cite{funk2006netflix,koren2009matrix} to Factorization Machines and their variants~\cite{rendle2010factorization,juan2016field}, ID embeddings have served as the primary carrier of collaborative signal across heterogeneous sparse features.

Deep learning has further scaled this paradigm. Embedding-and-MLP architectures~\cite{covington2016deep,cheng2016wide,guo2017deepfm,wang2017deep} replaced hand-crafted cross features with learnable interactions; sequential models~\cite{zhou2018deep,zhou2019deep,kang2018self,sun2019bert4rec} brought attention and recurrence to user-behavior sequences; and large recommendation models (LRMs)~\cite{naumov2019deep,zhai2024actions,zhang2024wukong,gui2023hiformer,zhang2026zenith} pushed model capacity onto massive ID-embedding tables. The bulk of model capacity thus resides in ID-embedding tables, and performance degrades sharply when the ID signal is insufficient. This limitation is particularly acute in livestreaming, where item lifetimes are extremely short.

\subsection{Multimodal Information in Ranking and Recommendation}

While effective for long-lived items, the ID-centric paradigm degrades under cold-start or short-lived conditions, where insufficient interactions prevent meaningful ID embeddings from forming---an effect especially pronounced in livestreaming, where room identifiers are ephemeral by nature. To address this, recent work has explored compact content-derived semantic codes such as YouTube's Semantic IDs~\cite{singh2024better} and TIGER~\cite{rajput2023recommender}, and, more broadly, has argued that sufficiently powerful modality encoders can match or surpass pure ID-based models in cold-start regimes~\cite{yuan2023go}. These developments point toward a broader trend of combining the memorization strength of IDs with the generalization capacity of content-derived representations.

Efforts to inject multimodal signals into industrial ranking systems can be broadly grouped into two families, distinguished by how tightly multimodal features are coupled with the ID embedding space: (i) \emph{dense multimodal features}, where frozen or lightly-trained embeddings are attached to the item as auxiliary inputs; and (ii) \emph{alignment and discretization}, where multimodal representations are explicitly aligned to, or compressed into, discrete codes more compatible with downstream ID-based interaction.

\noindent\textbf{Dense multimodal features.}
A common industrial recipe attaches frozen multimodal embeddings to items as auxiliary dense inputs to downstream rankers~\cite{sheng2024taobao,luo2025qarm}. The approach is simple but suffers from two well-documented limitations~\cite{luo2025qarm,deng2024em3}: \emph{representation mismatch} with the user--item interaction signal, and \emph{representation unlearning}, as frozen features cannot adapt to drifting preferences or business semantics.

\noindent\textbf{Alignment and discretization.}
To close this gap, recent work either aligns multimodal embeddings to the interaction space~\cite{liu2024alignrec,deng2024em3,abrec2025}, or discretizes them into ID-like semantic codes via quantization~\cite{luo2025qarm,ye2025dual,liu2025larm,zhang2024notellm2,zheng2025semid}; we adopt the prefix n-gram parameterization of Zheng et al.~\cite{zheng2025semid} in our embedding lookup (Section~\ref{sec:prefix_ngram}). Despite their diversity, all these systems share a common pattern: the multimodal signal serves as a \emph{complementary} feature alongside the dominant item ID in the ranker.

\noindent\textbf{Our positioning.}
In contrast, FLUID \emph{fully retires} the candidate-side item ID and lets LUCID serve as the sole content identifier on that side. We further train a dedicated multimodal encoder with cross-domain transfer from short videos to live streams, and target the extreme cold-start regime of live streaming (median room lifetime $\sim$40 min). Table~\ref{tab:mm_comparison} places FLUID against representative industrial and academic multimodal recommendation systems along four design axes. Two of these axes are uniquely satisfied by FLUID: it is the only method that (a)~\emph{retires} the candidate-side item ID rather than letting the multimodal signal coexist with it, and (b)~trains a multimodal encoder \emph{jointly across content domains} (short videos and live streams). Combined with \emph{Late} fusion and coverage of \emph{ephemeral} items, FLUID occupies a design cell that no prior industrial system has jointly occupied.

\begin{table*}[!htbp]
\centering
\caption{Positioning of FLUID against representative industrial and academic multimodal recommendation systems.}
\label{tab:mm_comparison}
\resizebox{\textwidth}{!}{%
\renewcommand{\arraystretch}{1.15}
\begin{tabular}{l|ccccccccc|c}
\toprule
 & Sheng et al.\! & AlignRec & QARM & EM3 & AB-Rec & DAS & LARM & NoteLLM-2 & Zheng et al.\! & \textbf{Ours} \\
 & \cite{sheng2024taobao} & \cite{liu2024alignrec} & \cite{luo2025qarm} & \cite{deng2024em3} & \cite{abrec2025} & \cite{ye2025dual} & \cite{liu2025larm} & \cite{zhang2024notellm2} & \cite{zheng2025semid} & \\
\midrule
\textit{Platform}        & Taobao    & Academic    & Kuaishou     & Kuaishou     & Academic     & Kuaishou     & Kuaishou     & Xiaohongshu  & Meta         & \textbf{Industry} \\
\textit{Year}            & 2024      & 2024        & 2025         & 2024         & 2025         & 2025         & 2025         & 2025         & 2025         & \textbf{2026} \\
\textit{Scenario}        & Ads       & Gen.        & SV           & EC           & SV           & Ads          & Live         & I2I          & Ads          & \textbf{Live} \\
\midrule
\textit{Fusion}                     & Early    & Early    & Early    & Early    & Early    & Early    & Early    & Late     & Late     & \textbf{Late} \\
\textit{Retires item ID}            & \ding{55} & \ding{55} & \ding{55} & \ding{55} & \ding{55} & \ding{55} & \ding{55} & \ding{55} & \ding{55} & \textbf{\ding{51}} \\
\textit{Ephemeral items}            & \ding{55} & \ding{55} & \ding{55} & \ding{55} & \ding{55} & \ding{55} & \ding{51} & \ding{55} & \ding{55} & \textbf{\ding{51}} \\
\textit{Cross-domain encoder}       & \ding{55} & \ding{55} & \ding{55} & \ding{55} & \ding{55} & \ding{55} & \ding{55} & \ding{55} & \ding{55} & \textbf{\ding{51}} \\
\bottomrule
\end{tabular}%
}
\end{table*}

\section{Method}
\autoref{fig:framework_overview} gives an overview of FLUID. The FLUID pipeline replaces the candidate-side item ID with content-derived multimodal codes in four stages: the \emph{cross-domain multimodal encoder} (\S\ref{sec:mm_model}) produces a content embedding for each live-stream slice; \emph{RQ-KMeans} (\S\ref{subsec:rq_kmeans}) discretizes the embedding into a hierarchical code we call \textbf{LUCID}; the \emph{prefix n-gram} scheme (\S\ref{sec:prefix_ngram}) maps each LUCID tuple into a learnable embedding; and the \emph{late-fusion ID-free ranker} (\S\ref{sec:late_fusion}) introduces LUCID as independent candidate-side tokens and retires the item ID via a staged warmup.
 
\subsection{Multimodal Encoder}
\label{sec:mm_model}
Live rooms are too short-lived for the ranker to learn useful per-item ID embeddings. FLUID therefore introduces a multimodal encoder as its first stage, producing a content-derived signal in place of the item ID. 
 
\subsubsection{Cross-domain Training Data}
Training the encoder requires pairs of queries and content slices labeled by user engagement---likes, shares, and watch-through. Many live rooms, however, end before accumulating enough engagement to form useful pairs. FLUID therefore trains a single encoder jointly on \emph{livestreams and short videos} in a shared embedding space, where the denser engagement signal from short videos improves generalization. 

\subsubsection{Architecture}
The multimodal encoder (\autoref{fig:framework_overview} top) consists of three components: a vision module, a text module, and a single-tower fusion module. The \textbf{Vision Module}, based on SigLIP2-base (native-resolution ViT)~\cite{tschannen2025siglip2} with a two-layer MLP projector, produces visual tokens $T_v \in \mathbb{R}^{N_v \times D_h}$. The \textbf{Text Module} tokenizes rich metadata---titles, OCR, ASR transcripts, author bios, audience comments, and sticker tags---into $T_t \in \mathbb{R}^{N_t \times D_h}$. The \textbf{Single-Tower Fusion Module}, built on Qwen3-Embedding-0.6B~\cite{zhang2025qwen3embedding}, processes the concatenated sequence $T = [T_v, T_t, \texttt{[EOS]}]$. Finally, the \texttt{[EOS]} hidden state is linearly projected to a 128-d embedding $z$. We adopt a single-tower architecture so that vision and text interact across the LLM's full depth, rather than only meeting at the output layer as in dual-tower designs. This single-tower advantage holds consistently across retrieval and classification benchmarks (Section~\ref{sec:ablation_mm}).

\subsubsection{Training Recipe}
The encoder is trained on a query-to-item (Q2I) contrastive task with InfoNCE loss~\cite{oord2018infonce} and false-negative masking. Queries are user search terms or MLLM-synthesized keywords; items are 2-minute content slices drawn from both livestreams and short videos, with positive pairs constructed from user-behavior signals (likes, shares, and watch-through). To preserve the pre-trained semantics in the LLM, we train in two stages: (1)~\textit{Alignment}---only the MLP projector and output projection are trained; (2)~\textit{Joint fine-tuning}---the full model is unfrozen and fine-tuned end-to-end at a reduced learning rate. Within each batch, we discard entire negative pairs whose query embeddings have pairwise similarity above a predefined threshold to reduce spurious negatives.


\subsection{Discrete Representation via RQ-KMeans}
\label{subsec:rq_kmeans}

Using the multimodal embedding $z$ corresponding to the 2-min content slice directly as a ranker input is suboptimal: embeddings from a generic vision--language objective are \emph{misaligned} with the user--item interaction signal, and a shared MLP over frozen multimodal embeddings lacks the expressiveness of embedding lookup tables of the ranker~\cite{luo2025qarm}. We therefore discretize $z$ via Residual Quantization K-Means (RQ-KMeans)~\cite{luo2025qarm,deng2025onerec} into an $L$-level codeword tuple we call \textbf{LUCID} (\textbf{L}ive \textbf{U}niversal \textbf{C}ontent \textbf{ID}entifier), which would later enter the ranker through a learnable embedding table co-trained with other sparse-ID features. We use RQ-KMeans rather than RQ-VAE because RQ-VAE collapses codebook entries under our online streaming retraining cadence, whereas K-means gives stable partitions once fit. We set $L{=}4$ and $N{=}64$. The resulting LUCID is a tuple $[c_1, \dots, c_4]$, e.g., $[33, 42, 63, 3]$.

Slice-level LUCID encodes 2-minute content dynamics---a streamer briefly switching from chatting to singing. A live room, however, has a persistent identity: the streamer's style, audience, and topical focus remain consistent throughout the broadcast, motivating a stable room-level identifier. We obtain this identifier by \emph{majority voting at each level}: at each quantization depth $l$, we take the most frequent codeword across all cumulative slices. Because residual quantization decouples coarse-to-fine semantics, per-level voting distills dominant content without mixing levels.


\subsection{Prefix N-gram LUCID Embedding}
\label{sec:prefix_ngram}

Each LUCID code---whether slice-level or room-level---is a tuple $[c_1, \dots, c_L]$ with $c_l \in \{0, 1, \dots, N{-}1\}$. We now describe the way to map the tuple to a learnable embedding $\mathbf{e}_{\text{LUCID}}$ consumed by the ranker backbone.

Earlier approaches~\cite{rajput2023recommender,luo2025qarm} use \textit{level-wise decoding}: Create $L$ independent embedding lookup tables $\mathbf{E}_1, \dots, \mathbf{E}_L$ with $\mathbf{E}_l \in \mathbb{R}^{N \times d}$, and let $\mathbf{e}_{\text{LUCID}}$ as the concatenation of $\mathbf{E}_l(c_l)$. 
However, in residual quantization, deeper levels encode refinements \emph{relative to the prefix path}: the same $c_l$ under different $[c_1, \dots, c_{l-1}]$ indexes entirely different semantic regions. For example, two slices with $c_1{=}0$ and $c_1{=}3$ that happen to share $c_2{=}2$ would both look up the same entry $\mathbf{E}_2(2)$ under level-wise decoding embedding, even though residual geometry indicates that those two sub-regions are unrelated.

We therefore adopt a \textit{prefix n-gram} embedding scheme~\cite{zheng2025semid}, which conditions each level's embedding on the full prefix path---analogous to the classical n-gram neural language model~\cite{bengio2003neural}, but with the context made explicit as a composite key:
\begin{equation}
    \bar{c}_l = \sum_{k=1}^{l} c_k \cdot N^{l-k}, \quad l = 1, \dots, L,
    \label{eq:prefix_index}
\end{equation}
where $\bar{c}_l$ uniquely identifies the path from the root to level $l$ in the quantization tree. With $N{=}4$, the two cases above yield different composite indices: $[c_1{=}0,\, c_2{=}2]$ gives $\bar{c}_2 = 2$ while $[c_1{=}3,\, c_2{=}2]$ gives $\bar{c}_2 = 14$---different slots in $\mathbf{E}_2$. Each level-$l$ table is expanded from $N$ to at most $N^{l}$ rows, and
\begin{equation}
    \mathbf{e}_{\text{LUCID}} = \mathbf{E}_1(\bar{c}_1) \,\|\, \mathbf{E}_2(\bar{c}_2) \,\|\, \cdots \,\|\, \mathbf{E}_L(\bar{c}_L).
    \label{eq:lucid_embed}
\end{equation}
The embedding at level $l$ is now a refinement of its own parent path, not a feature shared across unrelated sub-trees.

\subsection{Late Fusion and Staged Warmup}
\label{sec:late_fusion}

The final step is to incorporate LUCID embeddings to the production ranker. Our ranker backbone is a transformer-like architecture  \cite{zhang2026zenith} based on token-level feature interaction, where different groups of feature embeddings are represented by different tokens. We now provide details on the incorporation of LUCID embeddings as separate feature tokens into our ranker. 

\subsubsection{Room-level and slice-level integration}
\label{subsec:room_slice_integration}
Slice-level LUCID describes the current 2-minute content segment and captures short-term changes in the livestream, such as a streamer switching from chatting to singing. Room-level LUCID is obtained by the majority-voting procedure in Section~\ref{subsec:rq_kmeans} and provides a more stable descriptor of the live room. The two granularities are complementary to each other: the slice token provides transient multimodal evidence, while the room token provides persistent candidate identity.

Both embeddings use the prefix n-gram embedding scheme in Eq.~\ref{eq:prefix_index}, but they are parameterized by separate embedding tables. Sharing the same embedding tables leaves stable room semantics and fast-changing slice semantics into the same parameter space, which empirically weakens their expressiveness. We report the shared-table ablation in Section~\ref{sec:ablation_emb}.

\subsubsection{Fusion with the existing item ID feature}
\label{sec:gid_norm}

Let $\mathbf{g}$ denote the existing item ID embedding of the candidate item, and let $\mathbf{s}$ and $\mathbf{r}$ denote the prefix-n-gram embeddings of slice-level and room-level LUCID, respectively. The fusion strategies between content-based embeddings and ID-based embeddings can be broadly divided into two categories: early fusion and late fusion. 

Early fusion first maps the item ID and LUCID embeddings into a single candidate representation $\mathbf{h}$. For example: 
\begin{gather}
    \mathbf{h}_{\text{replace}} = \mathbf{s}, \\
    \mathbf{h}_{\text{concat}} = \mathbf{W}[\mathbf{g}\,\|\,\mathbf{s}], \quad \mathbf{W}\text{ a learnable projection}, \\
    \mathbf{h}_{\text{gate}} = \alpha \mathbf{g} + (1-\alpha)\mathbf{s}, \quad
    \alpha = \sigma(f(\mathbf{u})).
\end{gather}
where $\mathbf{u}$ can be the item ID embedding itself or side information such as exposure counters and item ID embedding norm. These formulations cover direct replacement, concatenation, rule-based gating, and learnable LARM-style gates~\cite{liu2025larm}. And the backbone treats the final representation as a single token:
\begin{equation}
    \mathcal{T}_{\text{early}} = \{\mathbf{h}\}.
\end{equation}
Even though the early fusion strategy sounds intuitive in that it combines two item-side information directly into a single token, as we shall demonstrate later in our experiments, it fails to extract the useful information from LUCID due to the strong memorization effect of the ID embedding. 

Late fusion instead leaves each signal as an independent token and lets the backbone learn their interactions:
\begin{equation}
    \mathcal{T}_{\text{late}} = \{\mathbf{g}, \mathbf{s}\}.
\end{equation}
This late fusion configuration allows us to fully leverage the strong interaction capacity of the ranker backbone to learn the incremental multimodal information from LUCID on top of the item ID. 

However, keeping the candidate-side item ID after the introduction of LUCID limits the generalizability ceiling of the ranker model, especially for livestreaming ranking where item ID embeddings are often under-trained because rooms are short-lived. This limitation is empirically validated in our later analysis in Section~\ref{sec:ablation_fusion}.
The final FLUID architecture therefore retires the candidate-side item ID completely and uses the room-level LUCID in its place, in addition to the slice-level LUCID. 
\begin{equation}
    \mathcal{T}_{\text{FLUID}} = \{\mathbf{r}, \mathbf{s}\}, \quad
    \mathbf{g} \notin \mathcal{T}_{\text{FLUID}} .
\end{equation}
This design relieves the model reliance on short-lived item ID embeddings while preserving two distinct content signals to characterize the candidate item: stable room identity and transient slice dynamics.

\subsubsection{Staged warmup in production}

\begin{figure}[!htbp]
    \centering
    \includegraphics[width=1\linewidth]{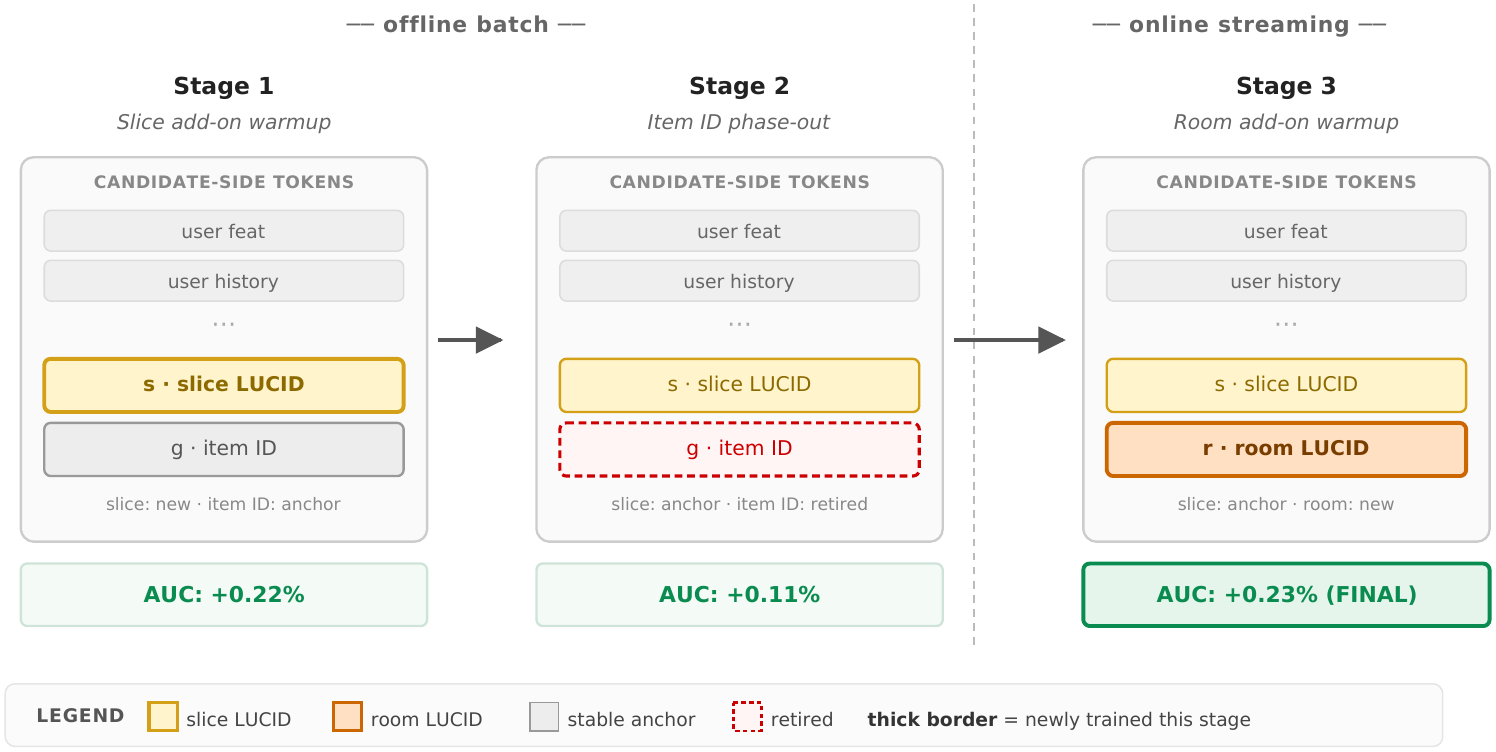}
    \caption{Three-stage warmup procedure for FLUID.}
    \label{fig:staged_warmup}
\end{figure}

Directly switching from the item-ID-based representation to $\mathcal{T}_{\text{FLUID}}$ is unstable because the LUCID embedding tables are newly introduced and the downstream interaction layers have been trained around an item-ID-bearing candidate representation. The deeper reason this transition cannot be done in a single retrain is an \emph{optimization asymmetry} between the item ID and LUCID. The ephemeral item ID embedding, although poorly generalizing, carries a strong item-level memorization signal that the backbone absorbs early in joint training, suppressing the slower but more generalizable LUCID signal; the resulting collapse is measured directly in Section~\ref{sec:ablation_fusion}. We therefore use a staged warmup procedure (Figure~\ref{fig:staged_warmup}) that changes only one item-side component per stage, leaving the rest as a stable anchor.

\begin{enumerate}[nosep]
\item \textbf{Slice add-on.} Initialize from a converged item-ID checkpoint and add slice LUCID as an independent token.
\item \textbf{Item ID phase-out.} Progressively zero out the candidate-side item ID, transitioning the model to rely on slice LUCID alone.
\item \textbf{Room add-on.} Initialize room-level LUCID tables from the warm slice-level LUCID embedding tables and train them independently to specialize on persistent room identity.
\end{enumerate}
After the third stage, we continue the production streaming online training schedule on the item-ID-free configuration with both slice-level and room-level LUCID. 

\section{Experiments}
\label{sec:experiments}

\subsection{Experiment Setup}
\label{sec:exp_setup}

FLUID comprises two trained models that we evaluate separately: the multimodal encoder of Section \ref{sec:mm_model}, which produces LUCID codes, and the ranking model of Section \ref{sec:late_fusion}, into which the LUCID codes are injected.

\noindent\textbf{Ranking model.}
All ranking experiments are conducted on the production ranking dataset of our industrial live-streaming platforms, which collectively serve a combined user base of over one billion globally. We adopt the full backbone, dataset, and training configuration of the production Zenith++ pipeline~\cite{zhang2026zenith} (dataset statistics in \autoref{tab:dataset_stats}), and on top of the Zenith++ backbone we plug in slice- and room-level LUCID codes produced by RQ-KMeans with $L=4$ residual levels and $N=64$ clusters per level.

\begin{table}[!htbp]
\centering
\caption{Statistics of the training dataset for the ranking model, shared with the production Zenith++ pipeline.}
\label{tab:dataset_stats}
\begin{tabular}{lccc}
\toprule
                                & \textbf{\# Instances} & \textbf{\# Features} & \textbf{\# Targets} \\
\midrule
Industrial Live Ranking          & 168B                  & 4{,}552               & 98 \\
\bottomrule
\end{tabular}
\end{table}

\noindent\textbf{Multimodal encoder.}
The encoder is trained on a query-to-item contrastive objective jointly over short videos and live streams from our industrial platforms, following the recipe in Section \ref{sec:mm_model}. We evaluate it on an internal benchmark suite covering retrieval (live and video keyphrase retrieval, live taxonomy classification, video e-commerce retrieval, live I2I retrieval) and semantic classification, under both linear-probing and full fine-tuning settings.

\subsection{Model Performance}
\label{sec:online_auc}


\textbf{Performance results.} \autoref{tab:model_perf} compares four ranker configurations on the CTR target. Adding slice LUCID as an independent token alongside the item ID improves AUC by $+0.22\%$, showing that multimodal semantics carry incremental signal. Directly removing the item ID (w/o item ID) is sharply negative, confirming the ranker's reliance on item ID memorization. Our FLUID design---retiring the item ID and replacing it with slice + room LUCID under late fusion and staged warmup (Section \ref{sec:late_fusion})---improves over the baseline on both metrics. It also surpasses the +Slice LUCID configuration by an additional $+0.01\%$ AUC and $-0.11\%$ logloss, showing that LUCID realizes its value as the item ID's successor rather than a supplement.


\begin{table}[!htbp]
\centering
\caption{Model performance across four configurations on the CTR target.}
\label{tab:model_perf}
\small
\setlength{\tabcolsep}{4pt}
\begin{tabular}{lcc}
\toprule
\textbf{Configuration} & \textbf{AUC} & \textbf{Logloss} \\
\midrule
baseline (with item ID)   & $0.7784$            & $0.1264$ \\
baseline + Slice LUCID    & $0.7801\,(+0.22\%)$ & $0.1263\,(-0.08\%)$ \\
baseline w/o item ID      & $0.7748\,(-0.47\%)$ & $0.1273\,(+0.65\%)$ \\
\makecell[l]{baseline w/o item ID \\ \quad + Slice \& Room LUCID \textbf{(FLUID)}}
                          & $\mathbf{0.7802\,(+0.23\%)}$ & $\mathbf{0.1262\,(-0.19\%)}$ \\
\bottomrule
\end{tabular}
\end{table}

\begin{table}[!htbp]
\centering
\caption{Online A/B test results. All values are relative changes vs.\ the production baseline. ``n.s.'' denotes changes not significant at $p<0.05$. Arms: \textbf{+Slice} = baseline + Slice LUCID; \textbf{$-$ID} = baseline w/o item ID; \textbf{FLUID} = baseline w/o item ID + Slice \& Room LUCID (FLUID).}
\label{tab:online_ab}
\begin{tabular}{lccc}
\toprule
\textbf{Metric} & \textbf{+Slice} & \textbf{$-$ID} & \textbf{FLUID} \\
\midrule
\multicolumn{4}{l}{\textit{Engagement quality}} \\
\quad Quality Watch Duration & $+0.43\%$ & $-0.02\%$ & $\mathbf{+0.55\%}$ \\
\quad Quality Watch Session  & $+0.39\%$ & $-0.10\%$ & $\mathbf{+0.51\%}$ \\
\midrule
\multicolumn{4}{l}{\textit{Cold-start, niche content, and diversity}} \\
\quad Cold-Start Room Views & $+1.15\%$ & $+1.58\%$ & $\mathbf{+2.05\%}$ \\
\quad Niche Room Views      & $+0.69\%$ & $+2.23\%$ & $\mathbf{+2.87\%}$ \\
\quad Unique Watched Tags   & $+0.55\%$ & $+0.20\%$ & $\mathbf{+1.63\%}$ \\
\midrule
\multicolumn{4}{l}{\textit{Retention}} \\
\quad Stay Duration & n.s. & $-0.05\%$ & $+0.07\%$ \\
\quad Active Hours  & n.s. & n.s.      & $+0.05\%$ \\
\bottomrule
\end{tabular}
\end{table}

\noindent\textbf{Online A/B test.}
\label{sec:online_ab}
To validate these offline gains in production, we further A/B-test the three arms (+Slice, $-$ID, FLUID) against the production baseline. \autoref{tab:online_ab} reports \emph{engagement quality}, \emph{cold-start, niche, and diversity}, and \emph{retention} metrics. \emph{Baseline + Slice LUCID} improves engagement and diversity but leaves retention unchanged, indicating that LUCID functions here as a secondary signal rather than a content identifier. \emph{Baseline w/o item ID} boosts cold-start and niche exposure---releasing long-tail traffic that the item ID had previously suppressed---but at a $-0.05\%$ cost in Stay Duration, showing that broader exposure trades off against ranking efficiency rather than coexisting with it. \emph{FLUID} is the only configuration delivering consistent gains across all three groups---engagement (Quality Watch Duration $+0.55\%$), diversity (Cold-Start Room Views $+2.05\%$), and retention (Stay Duration $+0.07\%$)---confirming that room-level LUCID truly \emph{replaces} the item ID on the candidate side, rather than merely \emph{supplementing} it.

\subsection{Ablation Studies}
\label{sec:ablation}

We conduct ablation experiments to validate the key design decisions in FLUID. Ablations on the multimodal encoder are reported on retrieval and classification benchmarks, while those on ranking components use CTR AUC unless otherwise stated.

\subsubsection{Multimodal Model}
\label{sec:ablation_mm}

\autoref{tab:ablation_training} ablates the three design choices behind this encoder: backbone (Qwen3-Embedding + SigLip2 vs.\ CLIP/Albert), fusion paradigm (single-tower vs.\ dual-tower with shallow fusion), and training schedule (two-stage alignment + joint fine-tuning vs.\ single-stage end-to-end). The backbone change contributes the largest gain (Live Q2I R@10: $43.96 \!\to\! 47.44$), with single-tower fusion and two-stage training each adding further improvements on every retrieval column.

\subsubsection{Cross-domain Case Review}
\begin{figure}[!htbp]
  \centering
  \begin{subfigure}[t]{0.49\linewidth}
    \centering
    \includegraphics[width=\linewidth]{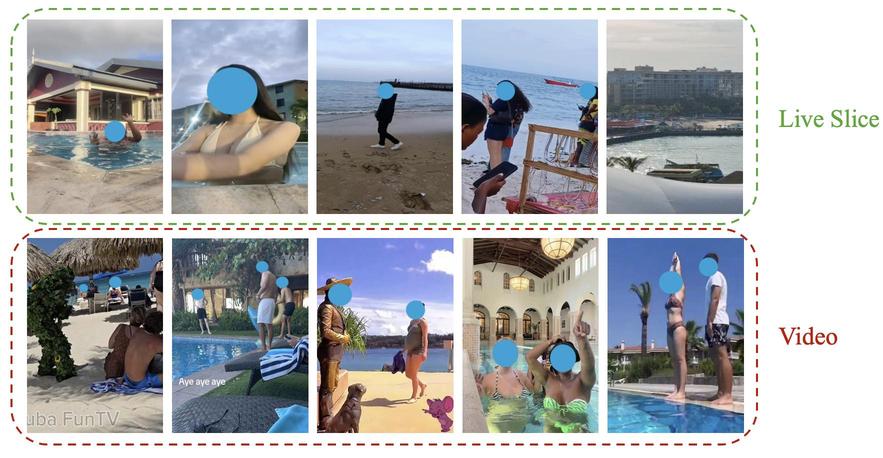}
    \caption{LUCID $(39, 41)$: ``swimming''.}
    \label{fig:rqcase1}
  \end{subfigure}\hfill
  \begin{subfigure}[t]{0.49\linewidth}
    \centering
    \includegraphics[width=\linewidth]{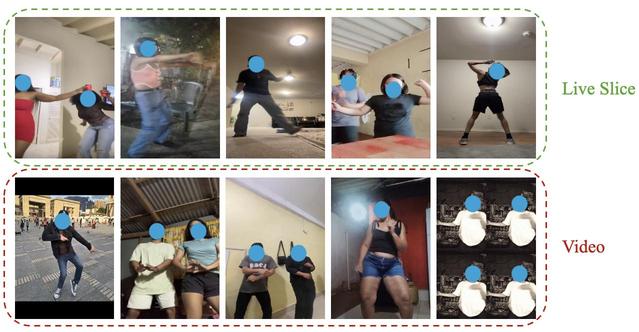}
    \caption{LUCID $(17, 26)$: ``dancing''.}
    \label{fig:rqcase3}
  \end{subfigure}
  \caption{Live slices and short videos grouped by LUCID: items sharing the same code remain semantically coherent.}
  \label{fig:rqcases}
\end{figure}

To illustrate the performance of our cross-domain encoder, we conduct a case review with two representative LUCID clusters in \autoref{fig:rqcases}: $(39, 41)$ for swimming, $(17, 26)$ for dance. Livestream slices and short videos remain semantically consistent under the same LUCID code, despite their genre differences.

\subsubsection{LUCID Embedding Design}
\label{sec:ablation_emb}

We ablate two design choices for mapping LUCID codes to embeddings on the candidate side: (i)~the \emph{embedding scheme}---Prefix n-gram (\autoref{sec:prefix_ngram}) vs.\ Level-wise decoding---and (ii)~whether slice- and room-level LUCID codes share a single embedding table or use separate ones (\autoref{subsec:room_slice_integration}). As summarized in \autoref{tab:emb_ablation}, both prefix n-gram embedding ($+0.11\%$ AUC over level-wise decoding) and independent slice/room tables ($+0.05\%$ AUC over the shared table) contribute positively, with prefix n-gram providing the larger gain.

\begin{table}[!htbp]
\centering
\caption{Ablation on LUCID embedding design.}
\label{tab:emb_ablation}
\begin{tabular}{lc}
\toprule
\textbf{Variant} & \textbf{AUC} \\
\midrule
Baseline (item ID only)                        & $0.7784$ \\
\midrule
\multicolumn{2}{l}{\textit{Embedding scheme}} \\
\quad Level-wise decoding                      & $0.7793\,(+0.12\%)$ \\
\quad Prefix n-gram (ours)                     & $\mathbf{0.7802\,(+0.23\%)}$ \\
\midrule
\multicolumn{2}{l}{\textit{Embedding tables}} \\
\quad Shared table                             & $0.7798\,(+0.18\%)$ \\
\quad Independent tables (ours)                & $\mathbf{0.7802\,(+0.23\%)}$ \\
\bottomrule
\end{tabular}
\end{table}

\begin{table*}[!h]
\caption{Multimodal encoder ablation results on backbone, fusion paradigm, and training schedule.}
\label{tab:ablation_training}
\centering
\begin{tabular}{l c c c c}
\toprule
\multirow{2}{*}{{Method}} & \multicolumn{2}{c}{{Live Keyphrase RET}} & \multicolumn{2}{c}{{Video Keyphrase RET}} \\
\cmidrule(lr){2-3} \cmidrule(lr){4-5}
& {Q2I R@10/50} & {I2Q R@10/50} & {Q2I R@10/50} & {I2Q R@10/50} \\
\midrule
\multicolumn{5}{l}{\textit{Backbone}} \\
\quad Baseline (CLIP-B/32 + Albert-V2) & 43.96/59.95 & 45.57/60.87 & 73.40/84.36 & 72.86/83.46 \\
\quad + Qwen3-Embedding & 46.15/60.43 & 47.18/61.32 & 84.74/91.97 & 85.30/91.40 \\
\quad \quad + SigLip2 ViT & \textbf{47.44/61.62} & \textbf{48.67/62.71} & \textbf{87.10/93.35} & \textbf{87.20/93.39} \\
\midrule
\multicolumn{5}{l}{\textit{Fusion paradigm}} \\
\quad Dual tower (shallow fusion) & 45.37/59.90 & 47.00/61.42 & 83.68/90.91 & 84.59/91.52 \\
\quad Single tower (ours) & \textbf{47.44/61.62} & \textbf{48.67/62.71} & \textbf{87.10/93.35} & \textbf{87.20/93.39} \\
\midrule
\multicolumn{5}{l}{\textit{Training schedule}} \\
\quad Single stage & 45.01/59.83 & 45.78/60.53 & 84.13/91.47 & 84.10/91.43 \\
\quad Two stage (ours) & \textbf{47.44/61.62} & \textbf{48.67/62.71} & \textbf{87.10/93.35} & \textbf{87.20/93.39} \\
\bottomrule
\end{tabular}
\end{table*}

\subsubsection{Candidate-side LUCID Integration: Fusion and Training Recipe}
\label{sec:ablation_fusion}

\begin{table*}[!h]
\centering
\caption{Ablation on candidate-side integration of LUCID with the existing item ID. $\Delta$ AUC is relative to the baseline (item ID only, AUC $=0.7784$).}
\label{tab:fusion_ablation}
\begin{tabular}{llcc}
\toprule
\textbf{Category} & \textbf{Method} & \textbf{AUC} & \textbf{$\Delta$ AUC} \\
\midrule
\multirow{6}{*}{Fusion mechanism}
& Early: Replace item ID with LUCID                       & $0.7774$ & $-0.13\%$ \\
& Early: Concat item ID + LUCID                           & $0.7785$ & $+0.01\%$ \\
& Early: LARM learnable gate~\cite{liu2025larm}           & $0.7783$ & $-0.01\%$ \\
& Early: LARM feature gate~\cite{liu2025larm}             & $0.7785$ & $+0.01\%$ \\
& Late: EM3 CIC alignment loss~\cite{deng2024em3}         & $0.7785$ & $+0.01\%$ \\
& Late: Independent token (= Stage 1 below)               & $0.7801$ & $+0.22\%$ \\
\midrule
\multirow{5}{*}{\makecell[l]{Training recipe\\}}
& Naive: joint training from scratch                       & $0.7784$ & $+0.00\%$ \\
& \multicolumn{1}{l}{\textbf{Staged warmup (ours, \S\ref{sec:late_fusion}):}} & & \\
& \quad Stage 1: Slice add-on (item ID + slice)            & $0.7801$ & $+0.22\%$ \\
& \quad Stage 2: Item ID phase-out (slice only)            & $0.7793$ & $+0.11\%$ \\
& \quad \textbf{Stage 3: Room add-on (slice + room, FLUID)} & $\mathbf{0.7802}$ & $\mathbf{+0.23\%}$ \\
\bottomrule
\end{tabular}
\end{table*}

This section ablates how LUCID is integrated into our token-based ranker backbone. The ablation spans two dimensions: (i)~the \emph{fusion mechanism}---whether slice LUCID and the item ID are merged before the backbone (early fusion) or enter as independent tokens (late fusion); and (ii)~the \emph{training recipe} under late fusion---how to train slice LUCID, progressively retire the item ID, and finally add room LUCID. \autoref{tab:fusion_ablation} organizes the resulting experiments.

\begin{figure}[!htbp]
    \centering
    \includegraphics[width=0.95\linewidth]{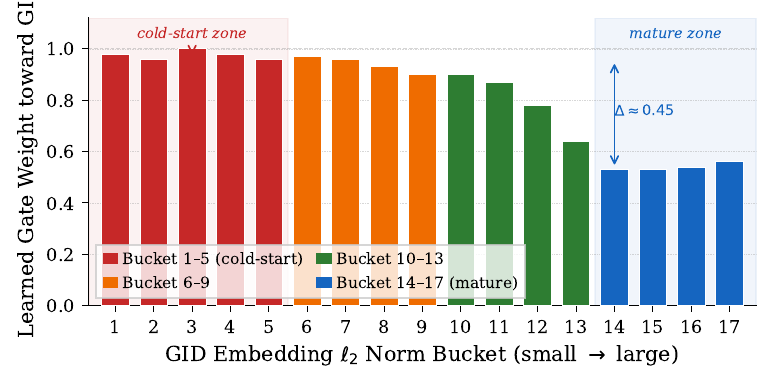}
    \caption{Gate inversion under the LARM learnable gate (\autoref{tab:fusion_ablation}). Converged gate weight on the item ID is plotted against item-ID embedding $\ell_2$ norm buckets (small $\to$ large). }
    \label{fig:gate_inversion}
\end{figure}

\noindent\textbf{Fusion mechanism.} We first ablate four early-fusion methods---parameter-free combinations (Replace, Concat) and learnable gates (LARM learnable, LARM feature)~\cite{liu2025larm}---and none of them yields meaningful improvement (max $+0.01\%$, min $-0.13\%$). \autoref{fig:gate_inversion} explains why for the LARM learnable gate: the converged gate weight collapses to the item ID, leaving slice LUCID essentially unused. We therefore turn to late fusion. EM3 CIC alignment~\cite{deng2024em3} adds an auxiliary loss that pulls the slice LUCID embedding toward the item ID embedding, but yields only $+0.01\%$. Removing the alignment loss and keeping slice LUCID and the item ID as plain independent tokens lifts AUC to $+0.22\%$. This configuration also serves as Stage~1 of our staged warmup below.

\noindent\textbf{Training recipe.} Naive joint training---where all embedding tables and the ranker backbone are reinitialized and trained from scratch---yields no improvement ($+0.00\%$): the ranker collapses to item-ID-only behavior, with slice and room LUCID failing to contribute. In our staged warmup (\S\ref{sec:late_fusion}), Stage~1 adds slice LUCID ($+0.22\%$), demonstrating that slice LUCID provides incremental signal beyond the item ID. Stage~2 removes the item ID ($+0.11\%$), revealing that slice LUCID alone cannot fully replace it. Stage~3 further adds room LUCID ($+0.23\%$ AUC), which fully compensates for the item-ID removal. After that, we try adding the item ID back, but it yields no additional AUC gain, confirming that LUCID has fully absorbed the candidate-side information previously provided by the item ID.

\section{Conclusion}

In this work, we present \textbf{FLUID}, the first framework to fully retire the candidate-side item ID from a production-scale livestreaming ranker. FLUID introduces a \emph{cross-domain} multimodal encoder that is jointly trained on short videos and livestreams to produce discrete semantic codes called \textbf{LUCID}. To adapt the ranker to LUCID, FLUID applies a staged warmup training scheme: it first leverages the ranker backbone for a late fusion of cold, slice-level LUCID embedding and the existing item ID embedding, and then replaces the item ID embedding with warm, room-level LUCID embedding before the final online incremental training. Deployed on our industrial livestreaming platforms with a combined user base of over one billion globally, FLUID delivers significant online gains of $+0.55\%$ Quality Watch Duration, $+2.05\%$ Cold-Start Room Views, and $+0.05\%$ Active Hours, covering engagement quality, cold-start exposure, content diversity, and user retention. This suggests a design lesson: when items are inherently short-lived, retiring the item ID is a more effective remedy than further fusion tricks for keeping the multimodal signal alive. We hope this work serves as a step toward content-grounded ranking in short-lived recommendation domains.

\bibliographystyle{ACM-Reference-Format}
\bibliography{ref}










\end{document}